\pdfoutput=1
\documentclass[11pt]{article}

 \usepackage[preprint]{acl}

\usepackage{times}
\usepackage{latexsym}
\usepackage{todonotes}
\usepackage{booktabs}
\usepackage{multicol}
\usepackage{siunitx}
\usepackage{adjustbox}
\usepackage{tabularx}
\usepackage{tikz}
\usetikzlibrary{positioning,fit,calc,decorations.pathreplacing}
\usepackage{float}
\usepackage{subfig}
\usepackage{subcaption}
\usepackage{multirow}

\usepackage[T1]{fontenc}
\usepackage[utf8]{inputenc}

\usepackage{microtype}

\usepackage{inconsolata}

\usepackage{listings}

\definecolor{javared}{rgb}{0.6,0,0} % for strings
\definecolor{javagreen}{rgb}{0.25,0.5,0.35} % comments
\definecolor{javapurple}{rgb}{0.5,0,0.35} % keywords
\definecolor{javadocblue}{rgb}{0.25,0.35,0.75} % javadoc
 
\title{\textsc{Fundus}: A Simple-to-Use News Scraper \\ Optimized for High Quality Extractions}

\author{Max Dallabetta \\ \href{mailto:max.dallabetta@hu-berlin.de}{\texttt{\small max.dallabetta@hu-berlin.de}}\And 
         Conrad Dobberstein \\ \href{mailto:conrad.dobberstein@informatik.hu-berlin.de}{\texttt{\small conrad.dobberstein@informatik.hu-berlin.de}}
         \AND
         Adrian Breiding \\ \href{mailto:adrian.johannes.breiding@hu-berlin.de}{\texttt{\small adrian.johannes.breiding@hu-berlin.de}}
         \And
         Alan Akbik \\ \href{mailto:alan.akbik@hu-berlin.de}{\texttt{\small alan.akbik@hu-berlin.de}}\\
         \AND \normalfont
         Humboldt Universität zu Berlin
}

\begin{document}
\maketitle
\begin{abstract}

This paper introduces \textsc{Fundus}, a user-friendly news scraper that enables users to obtain millions of high-quality news articles with just a few lines of code. Unlike existing news scrapers, we use manually crafted, bespoke content extractors that are specifically tailored to the formatting guidelines of each supported online newspaper. This allows us to optimize our scraping for quality such that retrieved news articles are textually complete and without HTML artifacts. Further, our framework combines both crawling (retrieving HTML from the web or large web archives) and content extraction into a single pipeline. By providing a unified interface for a predefined collection of newspapers, we aim to make \textsc{Fundus} broadly usable even for non-technical users. 
This paper gives an overview of the framework, discusses our design choices, and presents a comparative evaluation against other popular news scrapers. Our evaluation shows that \textsc{Fundus} yields significantly higher quality extractions (complete and artifact-free news articles) than prior work.

%Creating news corpora for subsequent NLP tasks poses significant challenges, particularly for researchers outside the realm of computer science, often necessitating extensive crawling and post-processing efforts. This paper introduces \textsc{Fundus}, a user-friendly news scraper leveraging the accessibility of Pythons' language and ecosystem to provide millions of high-quality news extractions with just a few lines of code. The framework combines a unified and streamlined interface with a predefined collection of newspapers, that can seamlessly transition between online and archive sources to effectively hide all web crawling complexities. Additionally, we prioritize quality through the use of hand-crafted extractors for each newspaper. 

The framework is available on GitHub under \url{https://github.com/flairNLP/fundus} and can be simply installed using \textit{pip}.
\end{abstract}

\section{Introduction and Motivation}

%Online news sites continuously report on a large range of topics (politics, technology, business, entertainment), mostly conveying information in the form of natural language text. As such, 

Online news articles are a favored data source for a wide-ranging set of NLP applications including social/political analysis ~\cite{Hamborg2019-12Autom-44511, masud2020hate, piskorski-etal-2023-semeval}, market prediction~\cite{10.5555/2832415.2832572, LI2020102212}, and are used as training data for language models~\cite{radford2019language,gururangan-etal-2022-whose}.

In such projects, it is often the first step to compile a corpus of news articles to analyze. This requires (1) identifying the URLs of news articles belonging to a particular set of online newspapers for download, and (2) extracting the article content from the surrounding HTML so that only the full article text remains. 

In particular the second task of \textit{content extraction} -- also referred to as web scraping or boilerplate removal~\cite{vogels2018web2text} -- is known to be challenging since each online newspaper uses different HTML and text formatting guidelines. This makes it non-trivial to distinguish between article content and other elements such as adverts, unrelated asides, captions, etc. 
%As for readability, we will refer to this task as content extraction throughout the paper.
To address these issues, several libraries have been developed that streamline the crawling and content extraction of online newspapers~\cite{Hamborg2017newsp-41887,10.1145/3366424.3383547,barbaresi-2021-trafilatura}.

\begin{figure}[t!]
    \centering
    \begin{tikzpicture}[
  node distance=3mm,
  nodes/.style={text width = 6mm},
  label/.style={anchor=west, align=left, xshift=5mm, font=\fontsize{6}{6}\ttfamily\selectfont},
]
    \node (title) [font=\fontsize{9}{9}\bfseries\selectfont, text width=4cm] {Oppenheimer wins Oscars };
    \node (author) [below=of title.west, font=\fontsize{7}{7}\selectfont, anchor=west, text width=4cm] {by Max Mustermann};
    \node (summary) [below=of author.west, font=\fontsize{8}{8}\itshape\selectfont, anchor=west, yshift=-2mm, text width = 4cm] { Oppenheimer wins 7 Oscars in 2024, including Best Picture. };
    \node (paragraphOne) [below=of summary.west, font=\fontsize{8}{8}\selectfont, anchor=west, yshift=-8mm, text width=4cm] { Christopher Nolan’s blockbuster "Oppenheimer" swept the Academy Awards in 2024, winning awards such as Best Picture, Best Actor and Best Director.
    };
    \node (subheadline) [below=of paragraphOne.west, font=\fontsize{8}{8}\bfseries\selectfont, anchor=west, yshift=-6.5mm, text width=4cm] {Nolan's first Oscar };
    \node (paragraphTwo) [below=of subheadline.west, font=\fontsize{8}{8}\selectfont, anchor=west, yshift=-2.5mm, text width=4cm] { This marks director Christopher Nolan's first [..] };
    \node (related) [below=of paragraphTwo.south, font=\fontsize{7}{7}\selectfont, anchor=center, minimum width=4cm] { Tags: Entertainment, Oscars 2024, ... };

    \node [draw=black!50, rectangle, rounded corners = 10, fit={(title) (author) (summary) (paragraphOne) (subheadline) (paragraphTwo) (related)}] {};

    \draw [decorate,decoration={brace,amplitude=4pt,mirror,raise=2pt},yshift=0pt] ($(summary.north west) + (-0.1,-0.1)$) -- ($(paragraphTwo.south west) + (-0.1,0.1)$) node [black,midway,xshift=-1mm,left=4pt,font=\fontsize{6}{6}\ttfamily\selectfont] {plaintext};

    % Adding labels as new nodes
     \node[label] (title_label) at (title.east) {title};
    \node[label] (author_label) at (author.east) {authors};
    \node[label] (summary_label) at (summary.east) {body.summary};
    \node[label] (paragraphOne_label) at (paragraphOne.east) {body.paragraph};
    \node[label] (subheadline_label) at (subheadline.east) {body.subheadline};
    \node[label] (paragraphTwo_label) at (paragraphTwo.east) {body.paragraph};
    \node[label] (related_label) at (related.east) {topics};

    % Drawing lines connecting labels to nodes
    \draw[-] (title_label.west) -- (title.east);
    \draw[-] (author_label.west) -- (author.east);
    \draw[-] (summary_label.west) -- (summary.east);
    \draw[-] (paragraphOne_label.west) -- (paragraphOne.east);
    \draw[-] (subheadline_label.west) -- (subheadline.east);
    \draw[-] (paragraphTwo_label.west) -- (paragraphTwo.east);
    \draw[-] (related_label.west) -- (related.east);

\end{tikzpicture}
    \caption{An example article scraped by \textsc{Fundus}. Next to the plain text of the article, attributes such as title, authors, paragraphs, subheadlines and topics are directly accessible.}
    \vspace{-2mm}
    \label{fig:attribute_overview}
\end{figure}

\begin{table*}[t!]
\vspace{-3mm}
    \centering
    \small
    \setlength{\tabcolsep}{4pt}
    \begin{tabular}{lllllll}
        \toprule
        \textbf{Library} & \textbf{[Info]}  &   \textbf{Language}  &  \textbf{Approach}  & \textbf{Extractor} & \textbf{Crawling} & \textbf{F1} \\
        \midrule
        \textsc{Fundus}  & (ours) & Python & Heuristics-based: Rules & bespoke & \textbf{yes} & \textbf{97.69}  \\
        \midrule
        Trafilatura &\cite{barbaresi-2021-trafilatura} & Python & Heuristics-based: Rules & generic &  \textbf{yes} & 89.81 \\
        BoilerNet & \cite{10.1145/3366424.3383547} & Python &   ML-based: Sequence labeling & generic &  no & 85.77 \\
        news-please &\cite{Hamborg2017newsp-41887} & Python &  Heuristics-based: Meta-rules & generic & \textbf{yes} & 85.81 \\
        jusText & \cite{Pomikalek2011thesis} & Python & Heuristics-based: Rules & generic &  no & 86.96 \\
        Boilerpipe & \cite{10.1145/1718487.1718542} & Java & ML-based: Node classification & generic &  no & 79.90 \\
        BTE & \cite{Finn2001FactOF} &  Python & Heuristics-based: Tag distributions  & generic & no & 87.14\\
        \bottomrule
    \end{tabular}
\vspace{-1mm}
    \caption{Comparison of \textsc{Fundus} to other prominent scraping libraries, some of which include crawling functionality. The F1-score measures the extraction quality on our benchmark, as detailed in Section~\ref{sec:evaluation}.
   % \todo[inline]{"accuracy of extraction"-> "extraction accuracy", etwas wordy. + Accuracy ist ja such selbst eine metric, vielleicht besser performance, correctness oder quality?}
   % \todo[inline]{boilerpipe ist im original in java https://github.com/kohlschutter/boilerpipe es gibt jedoch auch python bindings https://github.com/slaveofcode/boilerpipe3}
    }
    \label{tab:scrapers}
\vspace{-3mm}
\end{table*}

\noindent
\textbf{Limitations.}
However, existing libraries rely on generic methods for content extraction, based either on heuristics or trained machine learning models. This allows them to be applied across an arbitrary number of online newspapers, but comes at a cost of extraction accuracy: the quality of the news article texts varies depending on how well the heuristics or learned rules capture the HTML formatting of a particular newspaper. 

For instance, the evaluation presented in this paper shows that existing frameworks encounter difficulties with at least one newspaper, resulting in F1-scores below 60\% for all articles retrieved from this source. This means that, due to their generic nature, existing libraries provide no guarantee and no means to ensure that scraped articles are textually complete and without artifacts. 

Put more plainly, it may be argued that existing libraries prioritize \textit{quantity} (i.e.~scaling across many newspapers) over \textit{quality} (i.e. high-quality extraction of complete article texts and meta-attributes). This may cause problems in use cases in which the overall quality of a news corpus is more important than its quantity~\cite{li2023textbooks,marion2023more}. 

%This may cause problems in use cases in which the quality of text data is more important than its quantity~\cite{li2023textbooks,marion2023more}. 

%depending on how well the heuristics or learned rules capture the HTML formatting of a particular news source.

%While this allows them to scale across an arbitrary number of online newspapers, the quality of the extracted text may vary widely from source to source~\cite{}, depending how well the heuristics or learned rules capture the HTML formatting of a particular news source. Thus, there is no guarantee that parsed news articles are textually complete and without extraction errors.  

%Most of these approaches make use of a simple but effective aspect of web pages: "There is an inherent locality of relevant content" \cite{10.1145/3366424.3383547} for most web pages, as relevant content tends to be authored and rendered together. While this approach allows existing libraries to easily scale across various news sources, 
%\todo[inline]{after all the build-up, the following main argument of the limitation reads way too shallow and diffuse. We need something really grippy, pointy, and strict here - max}it cannot account for all publisher-specific layout variations.\\

%\noindent
%Depending on the library, varying amounts of user input are required. These can range from gathering the publisher domains up-to downloading all relevant HTML files and passing them into the framework.

\noindent
%\textbf{Fundus: A News Scraper Optimized for High Quality Extractions.} 
\textbf{Contributions.} 
With this paper, we present \textsc{Fundus}, a news crawling library in which we pursue an orthogonal approach to prior work. Rather than aiming for a set of general rules applicable to all newspapers, our library uses separate, manually created HTML content extractors -- referred to as \textit{parsers} within the library -- for each online newspaper. This allows us to match extraction methodologies specifically to a newspaper and thus manually optimize the accuracy of text extraction.

Further, as Figure~\ref{fig:attribute_overview} illustrates, this enables us to write more complex content extractors compared to prior work to preserve a news article's structure (distinguishing between paragraphs, sub-headlines, and the article summary), and extract meta-attributes such as topics. In more detail, our contributions are:

\begin{enumerate}
    \item We present the \textsc{Fundus} library, illustrate its ease of use, and discuss the merits and drawbacks of pursuing an approach of manually crafted, bespoke extractors for selected online newspapers. 
    
    \item We illustrate how \textsc{Fundus} can be used not only for news articles that are currently available online, but also scrape the extensive \textsc{CommonCrawl} web archive CC-NEWS. This allows users to create very large, high quality news corpora with only a few lines of code.  
    
    % also news articles from the extensive CommonCrawl news archive\textsc{CommonCrawl} news web dump\footnote{\href{https://commoncrawl.org/blog/news-dataset-available}{https://commoncrawl.org/blog/news-dataset-available}}, enabling the creation of very large news corpora with only a few lines of code.   
   %  \todo[inline]{- Sind die CC-News artikel nicht auch online? Vielleicht eher: "We illustrate how \textsc{Fundus} is not limited to processing continuous/up-to-date news articles but also news articles from the extensive CommonCrawl news archive/repository.", Archive oder repository statt dump würde denke ich auch nochmal hervorheben, dass es sich um vergangene news articles handelt. CommonCrawl nennt sich auch selbst Repository auf der main website. "Common Crawl maintains a free, open repository of web crawl data that can be used by anyone."
  %   - Ist es sinnvoll hier von CommonCrawl statt CCNews zu sprechen? In der Library hatten wir dies anders genannt.}
  
     \item We comparatively evaluate \textsc{Fundus} against well-known crawling/content extraction frameworks using a newly created dataset of paragraph-wise annotated HTML files, and provide statistics on its data potential leveraging \textsc{CC-NEWS}.     
 \end{enumerate}

We find that \textsc{Fundus} outperforms all other libraries in terms of yielding complete and artifact-free text (see Table~\ref{tab:scrapers}), thus indicating its usefulness for projects in which textual quality is a priority. To enable the NLP community to use \textsc{Fundus} in their projects -- and add parsers for new newspapers -- we open source \textsc{Fundus} under an MIT licence\footnote{Available at: \url{https://github.com/flairNLP/fundus}}. 

\begin{figure*}[t!]
\vspace{-3mm}
\noindent\begin{minipage}{.47\textwidth}
\begin{lstlisting}[caption=Crawl US-based publishers,label={lst:example_1},frame=tlrb]{Name}
# crawl US-based publishers
crawler = Crawler(PublisherCollection.us)

# crawl 10 news articles
articles = crawler.crawl(max_articles=10)

# print them out one-by-one
for article in articles:
    print(article)
\end{lstlisting}
\end{minipage}\hfill
\begin{minipage}{.47\textwidth}
\begin{lstlisting}[caption=Crawl one German publisher,label={lst:example_2},frame=tlrb]{Name}
# crawl one specific German publisher
crawler = Crawler(PublisherCollection.de.DW)

# crawl 10 news articles
articles = crawler.crawl(max_articles=10)

# print them out one-by-one
for article in articles:
    print(article)
\end{lstlisting}
\end{minipage}
\vspace{-3mm}
\caption{Two example usages of \textsc{Fundus} to crawl articles from (1) all supported US-based publishers, and (2)~only one specific German publisher ("Deutsche Welle").}
\label{fig:simple_examples}
\vspace{-3mm}
\end{figure*}
%\todo[inline]{the rest reads very well! - max}

%We evaluate \textsc{Fundus} in comparison to other well-known frameworks using a new dataset containing manually, paragraph-wise annotated HTML files. Scoring well on that evaluation leaves \textsc{Fundus} only limited by the reliance on manually implemented publishers. 

%But given its Open-Source nature, with a growing community, any gaps in publisher-coverage should be able to be closed, given enough time. Most importantly, this provides a degree of flexibility, that generic approaches cannot provide.

% In more detail, our contributions are:
% \begin{enumerate}
%     \item Release as open source-library
%     \item Experimental comparison against other libraries
%     \item Evaluation dataset
% \end{enumerate}
%The first step in such use cases is to crawl news data from the web. 

\section{Related Work}

Table~\ref{tab:scrapers} gives an overview of popular scraping libraries and a comparison to \textsc{Fundus}. 
%Existing approaches use either heuristics or ML-models for content extraction. Few approaches also include crawling functionality. 

\subsection{Content Extraction}
Existing approaches for content extraction are based either on heuristics or machine learning: 

\noindent 
\textbf{Heuristics-based extraction.} Early heuristic methods used the assumption that fewer HTML tags are used in main text content than in other elements. 
%, resulting in a distribution difference that can be detected as text or word density. 
For instance, \textsc{BTE} \cite{Finn2001FactOF} employs a cumulative tag distribution to identify the region with the lowest tag-to-text ratio as the main content.
%Subsequently, it extracts the corresponding text as main content. 
\textsc{jusText} \cite{Pomikalek2011thesis} segments HTML into content blocks based on selected tag types. Thereafter, these blocks are evaluated as content and distinguished from boilerplate using metrics such as link density, block length and block complexity.

%Creating a news corpus consists of new consecutive tasks, each of them posing unique challenges:
%\begin{itemize}
%    \item Discovering the web politely, while gathering pages at scale, and filtering for those relevant to the corpus (\textbf{crawling})
%    \item Detect and extracted main content from gathered pages (\textbf{scraping})
%\end{itemize}

%While there are several large web corpora built on news sites \cite{10.1145/3340531.3412762}, in need of specific outlets, topics, or date ranges, researchers are often required to create their own research corpus. Aggregating such a corpus consists of two 

%As for main content extraction, existing frameworks can be categorized into heuristic and machine learning-based approaches. Early \textbf{Heuristic} methods evolved around the assumption that HTML tags are less present in main content blocks, resulting in a distribution difference that can be detected as text or word density.\newline

%\textbf{BTE} \cite{Finn2001FactOF} employs a cumulative tag distribution to identify the largest "plateau", indicating the region with the lowest tag-to-text ratio. Subsequently, it extracts the corresponding text as main content.\newline

%\textbf{jusText} \cite{Pomikalek2011thesis} segments HTML into content blocks based on selected tag types. Thereafter, these blocks are evaluated as boilerplate employing metrics such as link or text density, block length or complexity, and the presence of stop words.\newline

%\noindent
Later approaches instead focus on the underlying DOM tree using a series of XPath expressions to determine regions of the tree as main content. For instance, \textsc{Trafilatura} \cite{barbaresi-2021-trafilatura} uses a cascade of XPath expressions to initially sanitize HTML content by removing unwanted sections and subsequently querying for relevant content. \textsc{news-please} \cite{Hamborg2017newsp-41887} facilitates a combination of state-of-the-art extractors.
%such as \textit{readability-lxml} and \textit{Newspaper3K}\footnote{\href{https://github.com/codelucas/newspaper}{https://github.com/codelucas/newspaper}}.

%Additionally, it integrates  \textit{readability-lxml}\footnote{\href{https://github.com/buriy/python-readability}{https://github.com/buriy/python-readability}} and \textit{jusText} as fallback mechanisms. 

%\textbf{news-please} \cite{Hamborg2017newsp-41887} facilitates a combination of state-of-the-art extractors, By default, it leverages the \textit{readability-lxml} and \textit{Newspaper3K}\footnote{\href{https://github.com/codelucas/newspaper}{https://github.com/codelucas/newspaper}} extractors. 

%\todo[inline]{Since we established the scraper terminology, i think that we should replace mentions of "extractor" with "scraper" - conrad}

\noindent
\textbf{ML-based extraction.} The second family of approaches formulates content extraction as a classification problem. 
%Early approaches classify blocks of text as content or boilerplate. 
For instance, \textsc{Boilerpipe} \cite{10.1145/1718487.1718542} uses decision trees to classify text blocks (uninterrupted text devoid of tags) as content or boilerplate. 
%It relies on structural, shallow text, and text density features.
\textsc{BoilerNet} \cite{10.1145/3366424.3383547} tokenizes web pages and trains a bidirectional LSTM to classify each segment.

%Each block is encoded into a vector representation, incorporating both the HTML tag path (from root to node) and textual content. Subsequently, bidirectional Long Short-Term Memory (LSTM) networks are employed to classify each segment as either content or boilerplate.

%\noindent
%Later ones consider page content as a sequence and formulate main content extraction as a sequence labeling problem.\newline

%Addressing this issue within those frameworks proves to be non-trivial, as their extraction methodologies are generic.

\noindent 
\textbf{Content extraction in \textsc{Fundus}.} Unlike prior work, we use bespoke extractors for each newspaper, thus allowing us to manually optimize for accuracy and attribute coverage. Although our approach inherently prioritizes quality, it also incurs a trade-off in terms of quantity, as it necessitates humans to create a separate extraction logic for each online newspaper. To manage this, we pursue a community-based approach and provide simple abstractions (and tutorials) to enable open source contributors to add support for new newspapers.  

%This naturally limits the scope of 

\subsection{Crawling}

Next to content extraction, identifying and downloading pages at scale can also be challenging. Such a system, which we refer to as a \textit{crawler}, should be "polite" (crawling only permissive online newspapers and keeping server workload low) and able to filter for pages relevant to the use case. However, the majority of existing libraries (see Table~\ref{tab:scrapers}) focus solely on content extraction, thus requiring users to resort to separate tools. 
%As shown in table \ref{tab:scrapers}, only a few integrate both crawling and content extraction functionalities.

\noindent 
\textbf{Crawling in \textsc{Fundus}.} We combine both crawling and content extraction in a single library. Unlike prior work which requires complex external configuration or comprehension of content maps like RSS feeds and sitemaps, \textsc{Fundus} provides pre-defined settings for each supported newspaper. In \textsc{Fundus}, users are only required to select a list of newspapers to include and issue a single method call, without additional configuration. This directly yields already extracted text. By hiding the underlying complexity, we aim to make \textsc{Fundus} broadly usable even for non-technical users.

\section{\textsc{Fundus}}

We introduce \textsc{Fundus} with a usage example, discuss our article and publisher-based logic, and illustrate how we distinguish between \textit{forward} and \textit{backward} crawling. 
%We also outline our content extraction mechanism.
%\todo[inline]{maybe emph forward and backward as other core terms}

%This section introduces \textsc{Fundus} with two simple usage examples and discusses the Article and Publisher classes. We also illustrate how \textsc{Fundus} supports and distinguishes between "forward" and "backward" crawling. Finally, we outline the abstractions used in our content extraction mechanism. 

%\todo[inline]{add part about tutorials etc.}

%\begin{figure*}[t]
%    \centering
%    \subfloat[\centering Crawl 10 news articles form US-based publishers]{{\includegraphics[width=7cm]{figures/fundus_example_1.png} }}%
%    \qquad
%    \subfloat[\centering Crawl one specific publisher for articles]{{\includegraphics[width=7cm]{figures/fundus_example_2.png} }}%
%    \caption{Two example usages of Fundus to crawl articles from (a) all supported US-based publishers, and (b) only one specific publisher.}%
%    \label{fig:example}%
%\end{figure*}

\subsection{Usage Example}\label{usage-examples}

We provide two example snippets on how to use \textsc{Fundus} to scrape news articles in Figure~\ref{fig:simple_examples}:
%, with Listings~\ref{lst:example_1} and~\ref{lst:example_2} respectively. 

\noindent 
\textbf{Listing~\ref{lst:example_1}: Crawl all US-based publishers.} This example demonstrates the process of scraping news articles from a selection of US-based publishers. First, we instantiate the \texttt{Crawler} object by passing \texttt{PublisherCollection.us} to it. This indicates that all US-based publishers currently supported in \textsc{Fundus} should be used as data sources. We then instruct the crawler to gather articles until a threshold of 10 articles is met (by passing \texttt{max\_articles=10}). This returns a generator\footnote{\textsc{Fundus} uses generators to prioritize responsiveness by delivering articles as they become available, rather than accumulating them for subsequent retrieval.} of \texttt{Article} objects, encapsulating the plain text of each news article alongside structured information such as the title, the author, and the date of crawling. Finally, we iterate over the generator, printing each article successively for human inspection.

\noindent 
\textbf{Listing~\ref{lst:example_2}: Crawl one specific source.} In the second example, we focus on articles from a particular publisher. We choose the German publisher DW ("Deutsche Welle") for this example. The code structure mirrors that of Listing 1, except that we instantiate the \texttt{Crawler} by passing \texttt{PublisherCollection.de.DW}. This narrows the search to a single publisher. 

%\textsc{Fundus} can be installed through pip ("\texttt{pip install fundus}"). After installation, users can crawl text by specifying which publishers to target, and how many articles to crawl. 

\lstdefinestyle{interfaces}{
  float=tp,
  floatplacement=tbp,
  abovecaptionskip=-5pt
}

\subsection{Articles} \label{articles}
\textsc{Fundus}' metadata and content extractions can be accessed through a single dataclass called \texttt{Article}. As indicated in the examples, users can obtain a quick overview of an article by simply printing it. This will output the article's title, a snippet of the extracted text content, the URL and publisher from which it was crawled, along with a timestamp indicating the publication time. 

All attributes for an \texttt{Article} are listed in Table~\ref{tab:attributes}. They are directly accessible using Python's dot notation. Attributes for each \texttt{Article} include its \textit{title}, textual \textit{body}, \textit{authors}, \textit{publishing\_date}, \textit{topics}, etc. The body attribute in particular captures the entire article structure including a summary, paragraphs and subheadlines, as depicted in Figure \ref{fig:attribute_overview}. 
%\todo[inline]{add reference to appendix?}

\begin{table}[t]
    \centering
    \small 
    \setlength{\tabcolsep}{4pt}
    \begin{tabular}{p{2cm}p{5cm}}
        \toprule
        \textbf{Attribute} & \textbf{Description} \\
        \midrule
        title & Title of the news article\\
        body & All article text with pararaph structure\\
        authors & Creators of the article\\
        publish\_date & Article release date\\
        topics & Publisher-assigned topics\\
        free\_access & Boolean indicating free accessibility\\
        ld & Parsed JSON+LD metadata\\
        meta & Parsed HTML metadata\\
        \midrule
        plaintext & Concatenated article body\\
        lang & Auto-detected article language\\ 
        html & HTML content and meta information\\
        exception & Indicates whether an exception occurred during extraction\\
        \bottomrule
    \end{tabular}
    \caption{Directly accessible attributes for each scraped \texttt{Article} (more details found in the Appendix, Table~\ref{tab:detailed_attributes}).
    }
    \label{tab:attributes}
\end{table}

\subsection{Publishers and Collections}

As the usage examples illustrate, users may specify which (set of) publishers to target when crawling for news articles. For \textsc{Fundus}, a publisher refers to an individual online newspaper, such as "Deutsche Welle". \textsc{Fundus} assumes that each online newspaper adheres to its own HTML and formatting guidelines. This means that for each publisher, we specify (1) where to find the URLs of each news article, and (2) how to extract the main textual content from downloaded HTML pages. This specification is created once (e.g. by a contributor to the \textsc{Fundus} repository) by creating a \texttt{Publisher}-specific enum object for the newly supported online newspaper.

Users can then pass this object to our \texttt{Crawler} to target this newspaper. To enhance accessibility and provide locality, we group publishers by their countries of origin within the \texttt{PublisherCollection}. This allows users to crawl all supported publishers of a specific country using their two-letter ISO 3166-2 language code.
%\footnote{\href{https://en.wikipedia.org/wiki/ISO_3166-2}{ISO 3166-2}}. 
We illustrate this by crawling all US-based publishers in Listing~\ref{lst:example_1}.
As of writing, the framework supports \textbf{39 publishers} spanning 5 different regions.

\subsection{Forward and Backward Crawling}

Internally, each \texttt{Publisher} defines one or multiple HTML sources, determining how a crawler locates the URLs of its news articles. Here, we distinguish between \textit{forward} and  \textit{backward} crawling.

\noindent 
\textbf{Forward crawling.} With forward crawling, we refer to accessing news articles that are currently available online on the news sites of supported newspapers. To identify URLs, we support the use of content maps like RSS feeds and sitemaps provided by the individual publisher. Sitemaps are typically exposed to crawlers via a "\texttt{robots.txt}" file, which also outlines user-agent-specific restrictions on subdomains and crawl intervals.

\noindent
\textbf{Backward crawling.} With backward crawling, we refer to accessing news articles in a static web dump. Specifically, we support the \textbf{CC-NEWS}\footnote{\url{https://commoncrawl.org/blog/news-dataset-available}} dataset provided by the \textsc{CommonCrawl} initiative. At the time of writing, this dataset comprises around 40 terabytes of WARC-formatted %\footnote{\url{https://iipc.github.io/warc-specifications/specifications/warc-format/warc-1.1-annotated/}} 
data, containing millions of news articles dating back to 2016. To handle such a large volume of data efficiently, \textsc{Fundus} offers the option to narrow crawls by a date range. Additionally, we stream WARC files and utilize the \textit{FastWARC} library \cite{bevendorff:2021} for in-memory processing to mitigate storage requirements.

% \noindent 
% \textbf{Backward crawling.} With "backward" crawling, we refer to accessing news articles in a static web dump. Specifically, we support the \textbf{CC-NEWS} dataset provided by the CommonCrawl initiative, comprising millions of news articles dating back to 2016. Due to the size of the dataset, we added the option to limit the search to a specific date range (for instance, to only new articles published in 2020).

% At the time of writing, the CC-News web dump consists of approximately 40 terabytes of WARC\footnote{\url{https://iipc.github.io/warc-specifications/specifications/warc-format/warc-1.1-annotated/}} data. To mitigate storage requirements when handling such a vast amount of data, we stream the WARC files from the CC-News dump and utilize the \textit{FastWARC} library \cite{bevendorff:2021} to process them in memory. To increase efficiency, \textsc{Fundus} leverages either concurrent or parallel computing methodologies depending on the task at hand. 

\begin{table*}[t!]
    \centering
     \vspace{-4mm}
    \small
    \setlength{\tabcolsep}{4pt}
    \begin{tabular}{lrrrrrrrrrrrrrrrrr}
        \toprule
        \textbf{Scraper} & \textbf{A} & \textbf{B} & \textbf{C} & \textbf{D} & \textbf{E} & \textbf{F} & \textbf{G} & \textbf{H} & \textbf{I} & \textbf{J} & \textbf{K} & \textbf{L} & \textbf{M} & \textbf{N} & \textbf{O} & \textbf{P}\\
        \midrule
        \textsc{Fundus} & 100 & 100 & 100 & 100 & 99 & 100 & 89 & 92 & 100 & 99 & 99 & 100 & 100 & 87 & 100 & 98\\
        \midrule
        BTE & 87 & 97 & 83 & 99 & 95 & 91 & 78 & 68 & 97 & \textbf{50} & 84 & 85 & 99 & 90 & 96 & 96\\
        jusText & 79 & 94 & 85 & 96 & 95 & 89 & 58 & 89 & 97 & \textbf{52} & 95 & 97 & 99 & 74 & 95 & 97\\
        Trafilatura & 93 & 99 & \textbf{42} & 99 & 97 & 94 & 84 & 97 & 100 & 74 & 100 & 95 & 100 & 67 & 97 & 100\\
        news-please & 100 & 91 & 93 & 100 & 81 & 95 & 78 & 97 & 97 & 82 & 85 & 85 & 71 & \textbf{32} & 100 & 85\\
        Boilerpipe & 82 & 96 & \textbf{5} & 97 & 75 & 93 & 75 & 96 & 96 & \textbf{52} & 75 & 95 & 88 & 77 & 91 & 87\\
        BoilerNet & \textbf{51} & 77 & 88 & 95 & 94 & 90 & 65 & 92 & 97 & 70 & 84 & 93 & 99 & 90 & 90 & 96\\
        \bottomrule
    \end{tabular}
     \vspace{-1mm}
    \caption{Rounded mean F1 scores of compared scrapers per publisher with scores below 60 highlighted. Publishers are: \textbf{A}: AP News; \textbf{B}: CNBC; \textbf{C}: Fox News; \textbf{D}: The Washington Free Beacon; \textbf{E}: The LA Times; \textbf{F}: Occupy Democrats; \textbf{G}: Reuters; \textbf{H}: The Gateway Pundit; \textbf{I}: The Guardian; \textbf{J}: The Independent; \textbf{K}: The Intercept; \textbf{L}: The Nation; \textbf{M}: The New Yorker; \textbf{N}: The Telegraph; \textbf{O}: The Washington Times; \textbf{P}: iNews}
\label{tab:scrapers_per_publisher}
\end{table*}

\subsection{Content Extraction}
The central component of \textsc{Fundus}' content extraction is the \texttt{Parser} class. It is individually implemented for every publisher and combines both generic and newspaper-specific extraction methods. The generic heuristics target structured information such as paywall restrictions, language detection, and meta-information (HTML tags and JSON+LD) and can be manually overwritten for specific publishers if necessary.

They are complemented with hand-tailored rules to extract the core parts of a news article such as the title, the textual body, and the authors. These rules are formulated as simple selectors (CSSSelect/XPath expressions) or metadata keys, and can typically be easily determined by inspecting the DOM tree of a few HTML examples. 
%We provide examples of such rules in Appendix~\ref{}.

Extraction rules are encapsulated as class methods for each parser and "registered" as attributes using a decorator. Each attribute in a parsed article can be directly accessed (c.f. Section~\ref{articles}).

\section{Evaluation}
\label{sec:evaluation}

We comparatively evaluate \textsc{Fundus} against prominent scraping libraries. Our goals are to (1) determine the quality of our bespoke content extraction approach compared to the generic approaches of prior work, and to (2) better understand the data potential of \textsc{Fundus}, i.e.~to estimate the size of news corpora that \textsc{Fundus} can create. 
%Further, we (3) measure the performance in terms of runtime of our forward and backward crawling approaches. 

%Since the main objectives vary between Forward and Backward Crawling, we have decided to evaluate the two cases separately.
%With the first, we want to show that due to \textsc{Fundus}' publisher-specific nature within the set of supported publishers, we outperform every other generic extraction framework. With the latter, we want to discover the data potential of WEB-Archives like CC-NEWS and show \textsc{Fundus}' capability in parsing archived pages.
%Last, we want to give a quick overview of the framework's performance regarding forward and backward crawling.

%\todo[inline]{Do we want to include simple XPATH as super baseline?}

% \begin{figure}
%     \centering
%     \includegraphics[width=\columnwidth]{latex/figures/complexity_boxplot.pdf}
%     \caption{Median and distribution of page complexity scores of our proposed dataset with and without paragraphs annotated as optional.
%     The complexity is defined as the ratio between HTML text tokens and tokens in the ground truth \cite{bevendorffEmpiricalComparisonWeb2023}. 
%     Whiskers mark \(1.5\times\) the interquartile range.
%     \todo[inlin]{Das sagt so wenig aus. Entweder andere Datensätze zum Vergleich mit einfügen, oder die Figure entfernen.}}
%     \label{fig:complexity}
% \end{figure}

%\todo[inline]{List objectives of evaluation: Determine quality, speed (?)}

\subsection{Experimental Setup}
\subsubsection{Evaluation Dataset}

To evaluate content extraction, we require a dataset of raw HTML pages and corresponding gold annotations of the journalistic content found on each page. This allows us to test whether content extraction libraries are capable of correctly distinguishing the article's text content from surrounding elements.
% To evaluate content extraction, we require a dataset of raw HTML pages and corresponding gold annotations of the plain text news article found on each page. This allows us to test whether content extraction libraries are capable of correctly distinguishing the news article text from surrounding elements.
Further, the dataset should cover the publishers in \textsc{Fundus}. As our survey of related work found no suitable datasets, we manually created our own\footnote{The dataset, scores, and evaluation metrics can be found at: \url{https://github.com/dobbersc/fundus-evaluation}}.

%We established two main criteria for a possible testing corpus. To get an accurate picture of the forward crawling functionality, the testing data is required to be up-to-date. Furthermore, a suitable dataset should cover a notable portion of our supported publishers. With these requirements, the only option we were left with, was to create our own corpus.
%\footnote{\href{https://github.com/flairNLP/fundus/tree/evaluation}{https://github.com/flairNLP/fundus/tree/evaluation}} 

\noindent 
\textbf{Data selection and annotation.} We select the 16 English-language publishers \textsc{Fundus} currently supports as the data source, and retrieve five articles for each publisher from the respective RSS feeds/sitemaps.
%\footnote{To ensure an accurate representation, we removed elements that did not contain any parsable text, e.g. videos.} 
We stress that the evaluation corpus consists only of articles that were published after the respective \textsc{Fundus} extractors were finalized. There is therefore no data contamination in our evaluation dataset. 

The selection process yielded an evaluation corpus of 80 news articles. From it, we manually extracted the plain text from each article and stored it together with information on the original paragraph structure. Annotation was separately performed by two authors of this publication. Our annotation guidelines can be found in Appendix~\ref{sec:guidelines} and include the option to mark individual paragraphs as "optional". To check for consistency between the two annotators, the first article of every publisher was annotated by both. Of 16 doubly annotated articles, 3 disagreements were discussed and resolved.   

{
\renewcommand{\arraystretch}{1.2}
\begin{table}[t!]
    \centering
    \setlength{\tabcolsep}{4pt}
    \begin{adjustbox}{width=\columnwidth}
    \begin{tabular}{llll}
        \toprule
        \textbf{Scraper} & \textbf{Precision} & \textbf{Recall} & \textbf{F1-Score} \\
        \midrule
        \textsc{Fundus} & \textbf{99.89}{\small±0.57\phantom{0}} & 96.75{\small±12.75} & \textbf{97.69}{\small±9.75\phantom{0}} \\
        Trafilatura & 90.54{\small±18.86} & 93.23{\small±23.81} & 89.81{\small±23.69} \\   
        BTE & 81.09{\small±19.41} & \textbf{98.23}{\small±8.61\phantom{0}} & 87.14{\small±15.48} \\  
        jusText & 86.51{\small±18.92} & 90.23{\small±20.61} & 86.96{\small±19.76} \\  
        news-please & 92.26{\small±12.40} & 86.38{\small±27.59} & 85.81{\small±23.29} \\
        BoilerNet & 84.73{\small±20.82} & 90.66{\small±21.05} & 85.77{\small±20.28} \\	
        Boilerpipe & 82.89{\small±20.65} & 82.11{\small±29.99} & 79.90{\small±25.86} \\
        \bottomrule
    \end{tabular}
    \end{adjustbox}
    \vspace{-1mm}
    \caption{Overall performance of \textsc{Fundus} and compared scrapers in terms of averaged ROUGE-LSum precision, recall and F1-score and their standard deviation.
    The table is sorted in descending order over the F1-score.}
    \label{tab:rouge_scores}
    \vspace{-2mm}
\end{table}
}

\begin{table*}
\vspace{-3mm}
    \centering
    \small
    \setlength{\tabcolsep}{3.2pt}
    \begin{tabular}{lrrrrrrrrrrrrrc}
        \toprule
        \textbf{Year} & \textbf{B} & \textbf{C} & \textbf{E} & \textbf{G} & \textbf{H} & \textbf{I} & \textbf{J} & \textbf{L} & \textbf{M} & \textbf{N} & \textbf{O} & \textbf{P} & \textbf{Total}\\
        \midrule
        2023 total & 19,628 & 75,363 & 40,048 & 63,664 & 12,951 & 55,899 & 176,913 & 2,380 & 2,973 & 57,600 & 15,388 & 28,911 & 551,718\\
        2023 body & 14,048 & 72,660 & 28,259 & 63,403 & 12,672 & 50,961 & 166,070 & 2,125 & 2,528 & 57,441 & 15,374 & 28,911 & 514,452\\
        \midrule
        2022 total & 21,820 & 209,452 & 40,531 & 115,811 & 0 & 96,504 & 217,238 & 2,296 & 4,904 & 61,053 & 19,947 & 30,285 & 819,841\\
        2022 body & 16,903 & 67,700 & 0 & 115,642 & 0 & 87,176 & 202,829 & 2,293 & 4,126 & 60,042 & 19,928 & 30,283 & 606,922\\
        \midrule
        2021 total & 26,741 & 101,906 & 47,019 & 248,619 & 40 & 93,954 & 112,498 & 2,345 & 4,652 & 73,953 & 45,184 & 20,791 & 777,702\\
        2021 body & 26,388 & 101,316 & 0 & 81,364 & 40 & 80,046 & 104,392 & 2,345 & 1,009 & 71,768 & 45,116 & 20,791 & 534,575\\
        \midrule
        2020 total & 31,725 & 109,155 & 54,901 & 399,925 & 33 & 97,174 & 0 & 2,839 & 5,318 & 89,393 & 90,065 & 71,070 & 951,598\\
        2020 body & 31,018 & 108,185 & 0 & 0 & 32 & 4,449 & 0 & 2,838 & 0 & 84,152 & 90,046 & 70,919 & 391,639\\
        \bottomrule
    \end{tabular}
\vspace{-1mm}
    \caption{Total number of articles extracted from \textbf{CC-NEWS} in the timeframe 01/01/2020 -- 01/01/2024, including a breakdown by online newspaper. Publisher identities correspond to those delineated in Table \ref{tab:scrapers_per_publisher}.}
    \label{tab:ccnews}
\vspace{-3mm}
\end{table*}

\subsubsection{Evaluation Metric}

We follow prior work  by \citet{bevendorffEmpiricalComparisonWeb2023} and use the ROUGE-LSum \cite{lin-2004-rouge} score, which is commonly used for evaluating the similarity between two text sequences, particularly in tasks such as machine translation. Here, we compare the extracted article text to the gold text. 

%Since the ground truth of our dataset is unstructured text rather than annotated HTML documents, we need to score the scraper performances via a text-based performance measure.
%Following previous work by \citet{bevendorffEmpiricalComparisonWeb2023}, we employ the ROUGE-LSum \cite{lin-2004-rouge} score, which is commonly used for evaluating the similarity between two text sequences, particularly in tasks such as machine translation.
%\todo[inline]{Do we need more details on ROUGE or an equation? A good summary and equation is in the bevendorff paper (better than the original rouge paper)}

For each article in the dataset, we calculate the precision, recall and F1-score using the ROUGE-LSum metric. This computation is performed with every possible combination of optional paragraphs removed from the ground truth, selecting the best F1 score from all options.
To determine the final score, we aggregate the scores of individual articles by computing the mean and the standard deviation. 

%We cap the number of combinations of removed optional paragraphs to \(2^{4}\) for computational reasons. For articles that exceed this threshold, we only compute the ROUGE-LSum score for the variant with all optional paragraphs included and without. Since consecutive sections of optional paragraphs often exhibit the same structure, we reason that the most significant impact on the final score is the entire presence or absence of optional paragraphs.
%\todo[inline]{Dieser teil ist ziemlich lang}

%Explain setup of experiments. Evaluation measures, baseline approaches (other libraries). 

%\todo[inline]{Not sure if we want to include a stripplot like this but this would illustrate the origin of the high standard deviation well and the underlying extraction distribution well. But takes more space...  for more see the key points in the discussion - conrad}

\subsection{Results and Discussion}

Table \ref{tab:rouge_scores} summarizes our findings. We make the following observations in regard to \textsc{Fundus}: 

\noindent 
\textbf{Highest overall F1-score.} We first note that our approach yields the highest quality extractions as measured by the ROUGE-LSum F1-score. This confirms our hypothesis that bespoke content extractors are naturally well-suited for high-quality text extraction. Further, this validates our assumption that publishers follow internally consistent formatting guidelines across all news articles.

% \todo[inline]{der obige paragraph enthält fundus in der überschrift. der darunter allerdings nicht direkt, um klarzustellen, dass die lower standard deviation sich such auf fundus bezieht. Vielleicht bei beiden Fundus weglassen?}

\noindent 
\textbf{Lower standard deviation.} We also note that \textsc{Fundus} has a lower variability of extraction quality -- as measured by the standard deviation -- than other approaches.
This indicates that our extractors are more consistent than generic approaches based on heuristics or on ML models. 
%\todo[inline]{Glaube ML wurde vorher als abkürzung nicht eingeführt}
We visualize this property in Table~\ref{fig:rouge_lsum_f1_score_stripplot} in the Appendix.

\noindent 
\textbf{Existing libraries struggle with at least one newspaper (Table~\ref{tab:scrapers_per_publisher}).} To get a better insight into the extraction capability of each compared library, we compute the F1-scores on a per-publisher basis. As Table~\ref{tab:scrapers_per_publisher} shows, we find that the F1-score widely varies from publisher to publisher for generic approaches, whereas \textsc{Fundus} yields consistent quality extractions.

\noindent 
\textbf{Errors remain.} However, we also note that despite manually-crafted, bespoke rules, our extraction is not perfect. Upon manual inspection, we find that a small portion of articles of a publisher deviates from standard formatting, for instance to emphasize quotations or include nested paragraphs. This particularly affected \textit{live news tickers} which some newspapers feature for selected events.

\subsection{Scalability}

Since \textsc{Fundus} is limited to a set of supported newspapers, a natural question is how much data one can expect to crawl using \textsc{Fundus}. 

\noindent 
\textbf{Data potential (Table~\ref{tab:ccnews}).}
To investigate, we extract news articles from the CC-NEWS web archive spanning the years 2020 to 2024. We find that 12 of our 16 English-language publishers are included in the archive. Further, despite crafting extraction rules targeting articles from 2023 onward, we note robust backward compatibility, with a significant decrease only noticeable in 2020 (e.g.~fewer URLs that yield text bodies). In total, we extract over 2 million articles with bodies. 

%\todo[inline]{Sollen wir den Table ins Appendix schieben? Wahrscheinlich schlecht für die Evaluation}

\noindent 
\textbf{Performance.} We evaluated the crawling performance of \textsc{Fundus} using a machine equipped with 2 Xeon 6254 CPUs, 756 GB of RAM, and a bandwidth of 10 Gbit/s. For CC-NEWS, we estimate the performance by focusing on the year 2023, as it constitutes the largest data dump among the four years evaluated. It comprises 201,586,338 unique URLs sourced from 34,229 different domains, resulting in approximately 8.2 terabytes of gzip-compressed WARC data. \textsc{Fundus} took 2.1~hours to yield the results presented in Table \ref{tab:ccnews}. 

In terms of forward crawling, we scraped 10,000 articles across all 39 supported publishers. Employing a delay of 1 second for subsequent calls on the same publisher, the process took 549 seconds.

\section{Conclusion and Outlook}

We presented \textsc{Fundus}, an easy-to-use news scraper built on the idea of bespoke content extractors for supported online newspapers. Our evaluation shows that our approach successfully optimizes for quality, indicating that \textsc{Fundus} is a viable option for use cases in which data quality is a priority. Further, we combine both crawling and scraping functionality in a single pipeline, and support access to the static web archive CC-NEWS.

With \textsc{Fundus}' open-source approach, we invite the community to contribute support for additional online newspapers. To assist in this process, we plan to investigate semi-automatic methods to suggest extraction rules in future work. 

%To manage this expanding set of publishers, plan to complement our tutorials with semi-automatic methods that suggest XPaths for

%plan to further extend the supervision of extraction quality already implemented within our GitHub repository.

%We show that with a few lines of code, users can create large, high quality corpora of news articles. 

%As comparable approaches spare hand-crafted rules in favor of scalability, we showcased that it is possible to maintain a set of handwritten extractors and optimize for quality while still being capable of scaling in size and creating large corpora thanks to web archives like CC-NEWS and extensive abstraction of the extraction process.

%With \textsc{Fundus}' open-source approach, we aim to enhance its extraction capabilities by adding new publishers and improving existing ones for better backward compatibility. To manage this expanding set of publishers, we plan to further extend the supervision of extraction quality already implemented within our GitHub repository.
% Bibliography entries for the entire Anthology, followed by custom entries
%\bibliography{anthology,custom}
% Custom bibliography entries only

\section*{Limitations}

The main limitation of our approach is its inherent lack of scalability across many online newspapers, since manual rules need to be written for each supported newspaper. As we argue in our paper, the benefits of extraction quality of our manual approach may outweigh quantity considerations depending on whether quality is a priority in an NLP use case. Additionally, though our approach does not easily scale across newspapers, it does scale across large web archives, meaning that we can retrieve large news corpora even with a limited number of supported publishers. Further, we aim to make it easy for the open source community to add support for new newspapers. 

A related limitation is that regular maintenance of extractions is necessary, since online newspapers might change their formatting guidelines over time. To monitor this, we automatically check whether \textsc{Fundus} is able to extract text content from currently online articles on a periodic basis. This flags whenever formatting guidelines have changed.  

%Further, to maintain backward compatibility while accommodating layout alterations, we have devised a class named \texttt{ParserProxy}. This class serves as a mediator between various parser versions tailored to the same newspaper, employing layout-adjusted extraction heuristics and organizing them according to date ranges. By providing the date an HTML page was crawled, calling a \texttt{ParserProxy} instance yields the appropriate parser version for extraction.
%this quality-quantity tradeoff 

%For \textsc{Fundus} regular maintenance is of the essence. This includes the constant verification that the extraction for each publisher works as intended, the occasional update, when it does not (e.g. due to layout changes, that can occur at any time), as well as the acquisition of more sources. In opposition to other libraries, \textsc{Fundus} regularly needs a certain level of attention to be able to continuously provide high-quality data and even more to continue expanding its availability. Hence, \textsc{Fundus} success is mainly limited by the manpower that is invested in it.

% A significant challenge in maintaining newspaper-specific and static content extraction lies in addressing layout changes over time. To mitigate this issue we designed a \texttt{ParserProxy} that proxies different parser versions of the same newspaper with layout-adapted extraction heuristics and mapping them to date ranges. By providing the date an HTML page was crawled, calling a \texttt{ParserProxy} instance yields the appropriate parser version for extraction.

\section*{Ethics Statement}
Newspapers play a pivotal role in modern society, often referred to as the fourth estate or fourth power. Maintaining independence necessitates self-financing for news media, thus evoking an inherent need for good-quality content to be adequately paid. However, the advent of Large Language Models (LLMs) revealed that web corpora, particularly news corpora, are often used for commercial benefit in a non-consensual manner. In response, our approach prioritizes the ethical acquisition of news articles by providing a simple option to crawl only those unrestricted by paywalls.

Moreover, we advocate for the non-commercial use of \textsc{Fundus}, aligning with our ethos of respecting intellectual property rights and promoting fair compensation for content creators. By fostering a culture of respect for intellectual property and fair compensation for content creators, we can help ensure the continued production of high-quality news and information for the benefit of society as a whole.

Another source of ethical concern stems from inherent biases in datasets obtained from the web~\cite{bender2021dangers}, as prior work has shown that language models trained over biased data tend to reflect these biases~\cite{haller2023opiniongpt}. With \textsc{Fundus}, users can specifically select which newspaper to include when creating a news corpus, thus giving some degree of control (for instance, over political biases) during corpus creation.

Lastly, also the datasets themselves are worthy of a discussion, as \textsc{Fundus} provides easy access to the CC-News dataset. The Common Crawl Foundation has measures in place to respect the resources and work of content creators and comply with the US Fair Use doctrine, which provides a legal basis. \citet{critical-analysis-common-crawl} and \citet{cc-opt-out-protocols} also illustrate limitations and biases of the Common Crawl dataset, which should be taken into account as well as acknowledging that not necessarily every rights-holder actively approved of their data being crawled and used.

\section*{Acknowledgements}
Max Dallabetta, Conrad Dobberstein and Alan Akbik are supported by the Deutsche Forschungsgemeinschaft (DFG, German Research Foundation) under Emmy Noether grant ``Eidetic Representations of Natural Language'' (project number 448414230). 
Alan Akbik is further supported under Germany’s Excellence Strategy "Science of Intelligence" (EXC 2002/1, project number 390523135).

\bibliography{custom}

\clearpage
\appendix

\section{Live Demo}
You can find a short live demonstration of our library on YouTube following this link: \url{https://youtu.be/9GJExMelhdI}

\section{Article Attributes}

\begin{table*}
    \centering
    \small
    \setlength{\tabcolsep}{2pt}
    \begin{tabular}{l l c c r}
        \toprule
        \textbf{Attribute} & \textbf{Description} & \textbf{Extraction} & \textbf{Methodology} & \textbf{Python type} \\
        \midrule
        \textbf{title} & Title of the news article & rule-based & metadata & str\\
        \textbf{body} & Object that allows direct access to paragraphs & rule-based & selectors & custom\\
        \textbf{authors} & Creators of the article & rule-based & mixed & list\\
        \textbf{publishing\_date} & Release date provided by the publisher & rule-based & mixed & datetime\\
        \textbf{topics} & Publisher-assigned topics & rule-based & mixed & list\\
        \textbf{free\_access} & Boolean indicating free accessibility & mixed & mixed & bool\\
        ld & JSON+LD data as extracted from the article & generic & selectors & custom\\
        meta & HTML meta tags as parsed from the article & generic & selectors & dict\\
        \midrule
        plaintext & Concatenated, stripped, and cleaned article body & - & - & str\\
        lang & Auto-detected article language & - & - & str\\ 
        html & contains raw HTML, origin URL, crawl date, and crawl source & - & - & custom\\
        exception & Exception indicating if an exception occurred during extraction & - & - & Exception\\
    \bottomrule
    \end{tabular}
    \caption{Article attributes alongside their description, extraction method, the applied methodology, and used Python type.}
\label{tab:detailed_attributes}
\end{table*}

Table \ref{tab:detailed_attributes} provides a comprehensive overview of all attributes of the \textsc{Fundus} Article class alongside additional information concerning the content extraction process, the methodology employed, and the Python data type utilized to represent each attribute internally.

\section{Annotation Guidelines}
\label{sec:guidelines}

For any given article we expect to extract the main textual content providing information on the article's topic which should align with editorial standards and be relevant to the headline. Additionally, relevant meta-information, e.g. declaration of third parties involved, additional information related, but not part of the main content, can also be extracted. Explicitly excluded are: 
\begin{itemize}
    \item The headline
    \item Captions of figures, images, and other objects
    \item Tables, due to the lack of a normalized representation
\end{itemize}
All extracted paragraphs are to be considered non-optional, unless one or more of the following conditions are fulfilled:
\begin{itemize}
    \item The paragraph's sole purpose is formatting 
    \item The paragraph is or is part of a summary of the article's contents
    \item The paragraph solely consists of meta-information (e.g. mentioning a contributing third party)
    \item The paragraph is not directly semantically related to the articles' content
\end{itemize}

\section{Standard Deviation of Compared Libraries}

\begin{figure}[H]
    \centering
    \includegraphics[width=\columnwidth]{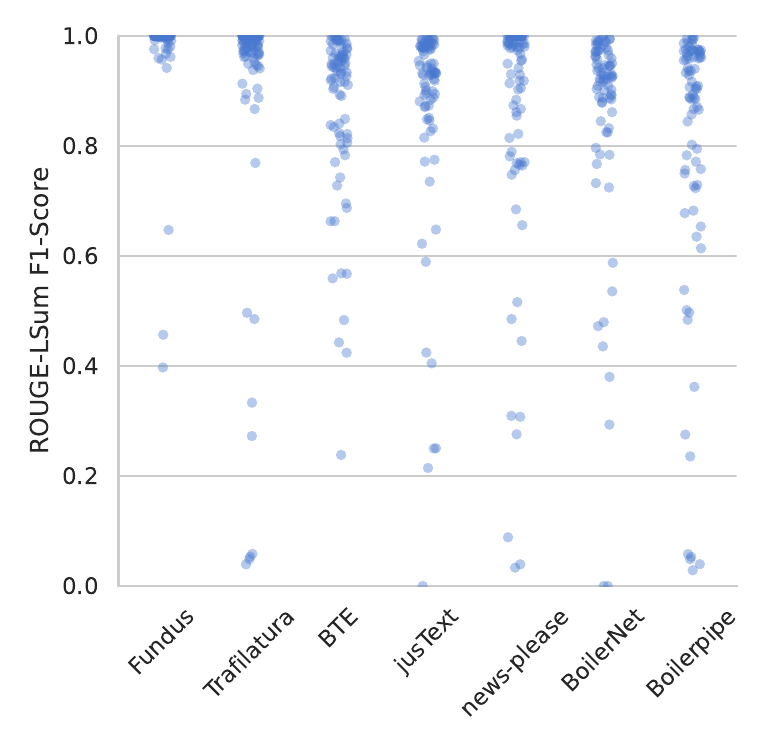}
    \caption{Distribution of ROUGE-LSum F1-scores of scraper extractions. The scrapers are sorted in descending order over the F1-score.}
    \label{fig:rouge_lsum_f1_score_stripplot}
\end{figure}

\section{CC-NEWS Crawling}
In addition to the examples outlined in Section \ref{usage-examples}, we aim to illustrate the ease of transitioning from forward to backward crawling. As depicted in Figure \ref{fig:commoncrawl}, this transition can be effortlessly achieved by substituting the employed crawler. Moreover, we offer a unified extraction interface, ensuring that switching between crawlers does not mandate parameter adjustments.

\begin{figure}[H]
\noindent\begin{minipage}{.47\textwidth}
\begin{lstlisting}[caption=Crawl US-based publishers,frame=tlrb]{Name}
crawler = Crawler(PublisherCollection.us)

# To retrieve articles from CC-NEWS instead one must simply exchange the pipeline
crawler = CCNewsCrawler(PublisherCollection.us)

for article in crawler.crawl(max_articles=100):
    print(article)
\end{lstlisting}
\end{minipage}\hfill
\caption{An example usage of Fundus to crawl articles from CC-NEWS}
\label{fig:commoncrawl}
\end{figure}

\end{document}